\pdfoutput=1

\documentclass[11pt]{article}

\usepackage[final]{acl}

\usepackage{times}
\usepackage{latexsym}
\usepackage{array}
\usepackage{booktabs}

\usepackage[T1]{fontenc}

\usepackage[utf8]{inputenc}

\usepackage{microtype}

\usepackage{inconsolata}

\usepackage{graphicx}

\title{Scaling Behavior for Large Language Models regarding Numeral Systems: An Example using Pythia}

\author{Zhejian Zhou$^{\diamondsuit}$ \quad Jiayu Wang$^\spadesuit$\quad Dahua Lin$^{\spadesuit\heartsuit}$\quad Kai Chen$^\spadesuit$\\
 $^{\diamondsuit}$ University of Southern California \quad $^\spadesuit$ Shanghai AI Laboratory \\ $^\heartsuit$ The Chinese University of Hong Kong\\
  \texttt{zhejianz@usc.edu} \quad \texttt{dhlin@ie.cuhk.edu.hk}\\
  \texttt{\{wangjiayu,chenkai\}@pjlab.org.cn}\\
}

\begin{document}
\maketitle
\begin{abstract}
Though Large Language Models (LLMs) have shown remarkable abilities in mathematics reasoning, they are still struggling with performing numeric operations accurately, such as addition and multiplication. 
Numbers can be tokenized into tokens in various ways by different LLMs and affect the numeric operations performance.
Currently, there are two representatives: 1) Tokenize into $1$-digit, and 2) Tokenize into $1\sim 3$ digit. The difference is roughly equivalent to using different numeral systems (namely base $10$ or base $10^{3}$). In light of this, we study the scaling behavior of different numeral systems in the context of transformer-based large language models. We empirically show that a base $10$ system is consistently more data-efficient than a base $10^{2}$ or $10^{3}$ system across training data scale, model sizes under from-scratch training settings, while different number systems have very similar fine-tuning performances. 
We attribute this to higher token frequencies of a base $10$ system.
Additionally, we reveal \textit{extrapolation} behavior patterns on addition and multiplication. We identify that base $100$ and base $1000$ systems struggle on token-level discernment and token-level operations. We also sheds light on the mechanism learnt by the models.
\end{abstract}

\section{Introduction}
Large Language Models (LLMs) have stormed the world with their amazing reasoning abilities \citep{gpt4,gemini,llama2}. 
However, numeric operations remain challenging for LLMs to comprehend under the architecture of Transformer \citep{attention,lee2023teaching,yuan2023arithben,rasp-l,mcleish2024transformers}. 
Several techniques have been proposed to improve the performance of numeric operations including improving positional embeddings \citep{kazemnejad2024impact, mcleish2024transformers} and using scratchpad \citep{nye2021show,goat}.
These works mostly focus on a random initialized Transformer with $1$-digit tokenization.
However, pre-trained LLMs have various tokenizers that can affect the numeric operations performances.
Currently, there are two main tokenization schemes: 1) Tokenize into $1$-digit \citep{llama,llama2,mistral,qwen,gemma,shao2024deepseekmath}, and 2) Tokenize into $1\sim 3$ digit \citep{pythia,gpt4,internlm2}. An example of different tokenization\footnotemark\  is shown in Table~\ref{tab:segmentation}.
Abstracting away practical details of tokenizers, these two schemes can be viewed as using a base $10$ numeral system versus a base $10^3$ system.
The former aligns better with human intuition and the prevalent base $10$ system in daily usage.
Yet, the latter encodes numbers into fewer tokens.
Our question follows intuitively: What is the difference between these schemes in numeric operations?

\begin{table*}[h!]
    \centering
    \small
    \begin{tabular}{lll}
        \toprule
        Type & Models & Tokenize 31415926535 \\
        \midrule
        $1$-digit & Llama1-2/Mistral/QWen/Gemma &  ['3', '1', '4', '1', '5', '9', '2', '6', '5', '3', '5'] \\
        \midrule
        Multiple & Pythia/GPT-4o/Llama3/InternLM & ['314', '159', '265', '35'] \\
        \bottomrule
    \end{tabular}
    \caption{Tokenizing 31415926535 via different large language models.}
    
    \label{tab:segmentation}
\end{table*}
We resort to data-scaling efficiency to answer this question. That is, there would be substantial differences in the scaling behavior of these numeral systems. 
Intuitively, a base $10$ system has a smaller set of tokens that could appear at each position. However, it would take up more context length to represent the numbers. 
Out of practical considerations, we choose to restrict our study to the base $10$, base $10^2$, and base $10^3$ systems which adhere to tokenizers of existing large language models. 

To design experiments for scaling behavior, we identify the following critical dimensions: 1) numeral system 2) data scale, and 3) model size. To further corroborate the generalizability of our claim, we also test if our conclusion holds for different numeric operations. On the other hand, it is possible that pre-trained models have a bias towards $2\sim3$ digit tokens. To strengthen our claim, we test if our observed trend holds irrespective of whether our models are trained from-scratch or fine-tuned. 

We observe that a base $10$ system is consistently more data-efficient when trained from-scratch, and that fine-tuned models perform comparably. We attribute this to higher token frequencies in base $10$ training data. We believe our observation could transfer to other tasks. For example, having a smaller state/action space could be favorable in terms of data efficiency for a sequential planning task. 

We also study the length extrapolation behaviors of different numeral systems. We identify that base $100$ and base $1000$ systems struggle on token-level discernment, and on learning token-level operations. We further shed light on the mechanisms learnt by models.

Our contributions can be summarized as follows:

\begin{itemize}
    \item We reveal that A base $10$ system is consistently more data-efficient than a base $10^2$ or base $10^3$ system under different data scales, model sizes, and different operators, especially training from-scratch.
    \item We showcase through extrapolation experiments that base $100$ and base $1000$ systems struggle on token-level discernment and on learning token-level operations.
    \item We identify several calculation patterns in the extrapolation setting. Such patterns include truncated addition and extrapolation of base-$10$ carry. 
  
\end{itemize}

\footnotetext{We provide a discussion on tokenization schemes in Appendix \ref{app:tokenization}.}
\section{Related Work}
\noindent \textbf{Numeral System} \quad A numeral system represents a number by a sequence of tokens within pre-defined sets. In order to perform numeric operations, the model would have learned to discern between the tokens precisely. Numeral systems are closely related to tokenizers. We first review prevalent tokenization conventions. 
Llama1/2 \citep{llama,llama2} tokenize numbers into $1$-digit, enforcing a base-$10$ system. 
This design is also adopted by other general-domain LLMs \citep{mistral,qwen,gemma} and math-specialized model Deepseek-Math \citep{shao2024deepseekmath}.
On the other hand, the most capable model to date, GPT-4o, tokenizes numbers into $1\sim 3$ digit, which is roughly equivalent to using a base $10^3$ system. 
To the best of our knowledge, no one has systematically studied how the numeral system affects the transformers' arithmetic ability.

\noindent \textbf{Arithmetic Operations in Transformers} \quad
To improve the arithmetic abilities of the transformer \citep{wang2021exploring,nogueira2021investigating}, people have designed positional embeddings \citep{kazemnejad2024impact,mcleish2024transformers}, scratchpad \citep{nye2021show}, and special training procedures \citep{goat,icot}. In this paper, we do not improve the performance of arithmetic operations, but aim to understand the scaling impact of choices of numeral systems. We focus on using decoder-only transformers to generate the results of arithmetic operations directly (i.e. without a scratchpad).

\noindent \textbf{Scaling Laws in Large Language Models} \quad Scaling laws\footnotemark \ have been widely studied in the context of LLMs \citep{kaplan2020scaling,hernandez2021scaling,gao2023scaling,bi2024deepseek} which aims to predict model losses based on different data scales and model parameters. Different from this line of research, we do not aim to accurately predict performances when we scale up computing. We leverage scaling behavior as a proxy to study the impact of numeral systems selection.

\footnotetext{We discuss the connections between our work and scaling laws in Appendix \ref{app:scaling_law}.}

\section{Scaling Behavior Experiment Designs}\label{sec:exp_design}
To understand how the numeral systems affect numeric operation in LLMs, we identify the following dimensions of interest for our experiments when training an LLM with numerical operation: 1) \textit{numeral system} 2) \textit{training data scale} 3) \textit{model size} 4) \textit{from-scratch} or \textit{fine-tuning} 5) \textit{different operations}. 
 
For 1) and 2), we generate synthetic inputs according to the process explained in section \ref{sec:data_gen}. For 3), we make use of the Pythia scaling suite \citep{pythia} for ranging over different model sizes. For 4), we replicate experiments for both settings to the best of our effort. For 5), we choose to include results of addition and multiplication. 

We list the complete configurations for our experiments. 1) numeral system: base $10$, base $10^2$, base $10^3$ 2) training data scale: $2^{13\sim 19}$ training samples 3) model size: 70M, 410M, 1.4B, 6.9B, 12B from Pythia 4) random-initialized or fine-tuned from Pythia (i.e. from-scratch or fine-tune) 5) operations\footnotemark: addition, multiplication.
After choosing a configuration, we train our model using our generated data and evaluate the model on a non-overlapping evaluation set.
The training procedure is the same as instruction-tuning a language model which masks the prompt (for example $12 + 23 = $) and only calculates the losses on the outputs ($35$).

\subsection{Synthetic Data Generation} \label{sec:data_gen}
We generate synthetic data of scales $2^{13\sim 19}$ for numeral systems base $10$, base $10^2$, and base $10^3$. 
We abstract away the nitty-gritty details involved in practical tokenization schemes and generate synthetic input ids and labels directly. 

We first illustrate our training distribution generation process using \textit{addition} as an example. Let \textit{a} and \textit{b} be two operands, each row would be in the form of $a + b = c$. Let \textit{la} and \textit{lb} be the digit lengths of \textit{a} and \textit{b} in base $10$. We fix $la, lb \in [1,10]$, and we attempt to evenly distribute generated data over $la\times lb$. If the total number of pairs for $la\times lb$ is smaller than we request, we take all possible pairs. No pairs are repeated during our generation.

\begin{figure}[t] 
  \includegraphics[width=\columnwidth]{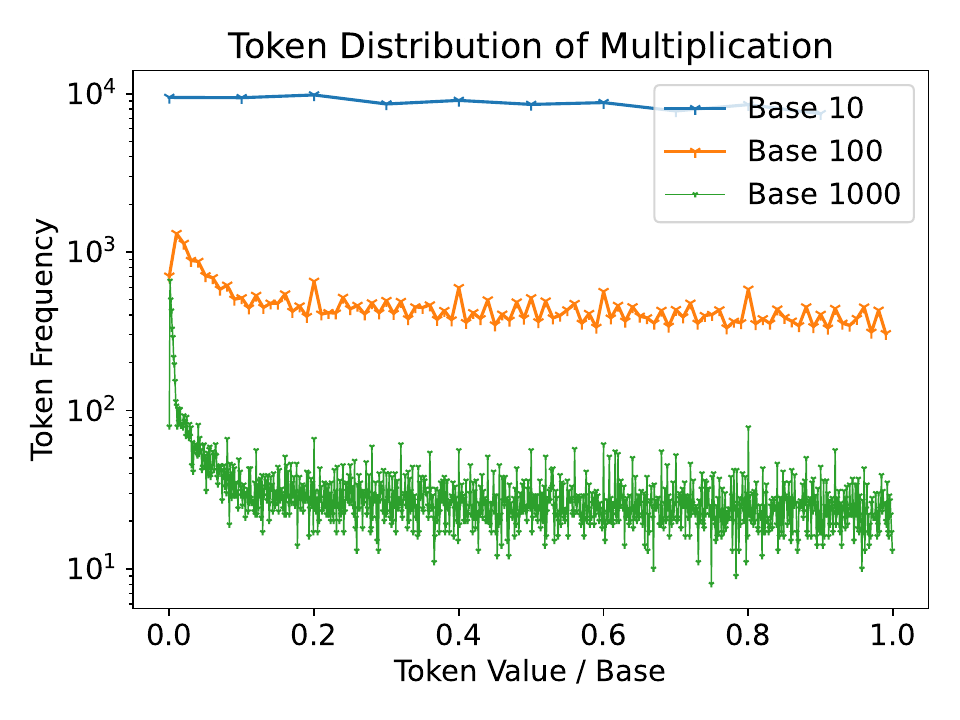}
  \caption {Answer Token Distribution for Multiplication. We sample $2^{13}$ addition samples to illustrate the distribution. Token values are normalized to $[0,1]$.}
  \label{fig:token_dis}
\end{figure}

Based on our generated training distribution, we convert each number into the corresponding base $10$, base $10^2$, and base $10^3$ representations. Note that this could be easily done by grouping digits in the original base $10$ representation. We then map the digit numbers onto their corresponding token ids. Intuitively, base $10$ would have 10 ids (corresponding to $0\sim 9$), base $10^2$ would have 100 ids, and base $10^3$ would have 1000 ids.
In Figure \ref{fig:token_dis}, we demonstrate the answer token distribution for each numeral system. 

Importantly, \textbf{token frequencies} of a base $10$ system are at least an order of magnitude larger than those of a base $100$ or base $1000$ system. \textbf{We believe that higher token frequencies lead to better trained models.} This is the reason to the superior performances of a base $10$ model.

We obtain the distribution by converting all answers into ids. We normalize the token values from each numeral system by dividing against the base. 
As our sampling result shows, the probability density gets more imbalanced as the base gets larger. For example, tokens $0\sim9$ are one magnitude more likely to appear, followed by two-digit tokens, then three digit tokens. Such a phenomenon could have deeper roots in number theory. In this paper, we accept this experiment fact and continue with our exploration. 

\footnotetext{We provide a more detailed discussion on task selection in Appendix \ref{app:task_choice}.}
\subsection{Evaluation Setup}
We sample non-overlapping operand pairs for evaluation. We attempt to evenly sample 1000 pairs for each $la \times lb$. If half of the total number of pairs is smaller than 1000, then we reserve half for evaluation. Overall, we strive to make sure that the training and evaluation sets are from the same distribution and have no overlap. 

\begin{figure*}[ht] 
  \includegraphics[width=\linewidth]{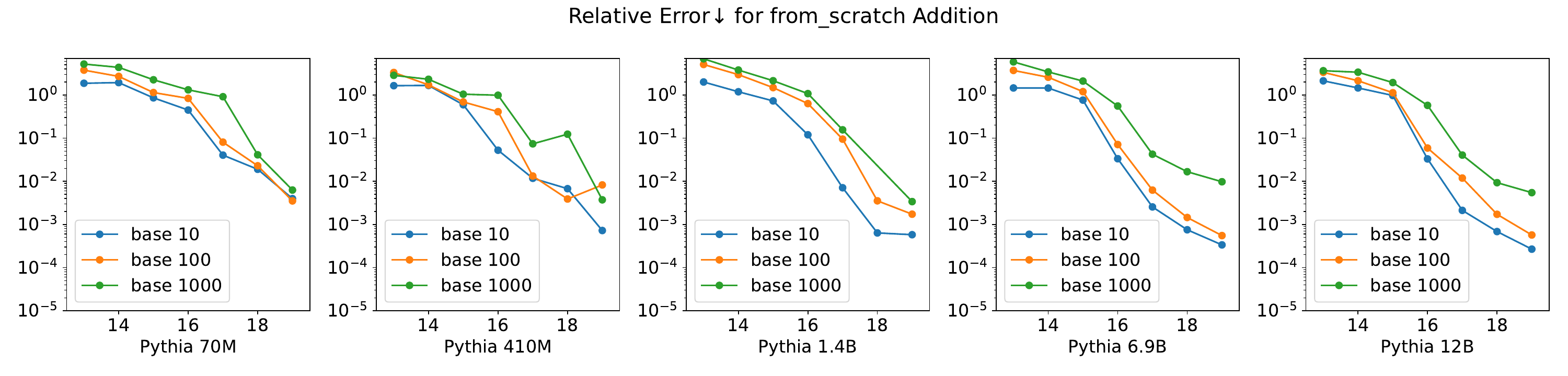}
  \includegraphics[width=\linewidth]{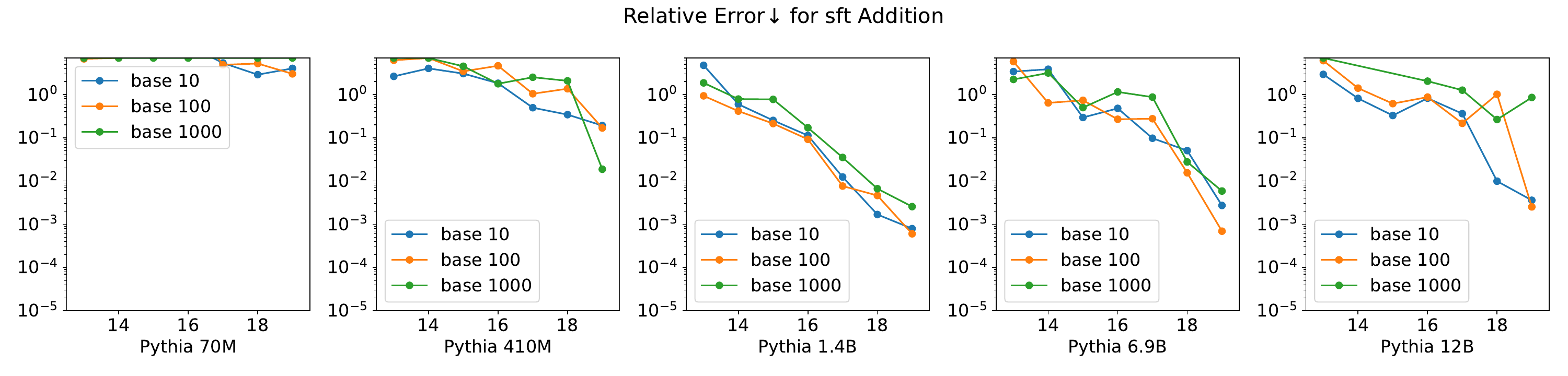}
  \includegraphics[width=\linewidth]{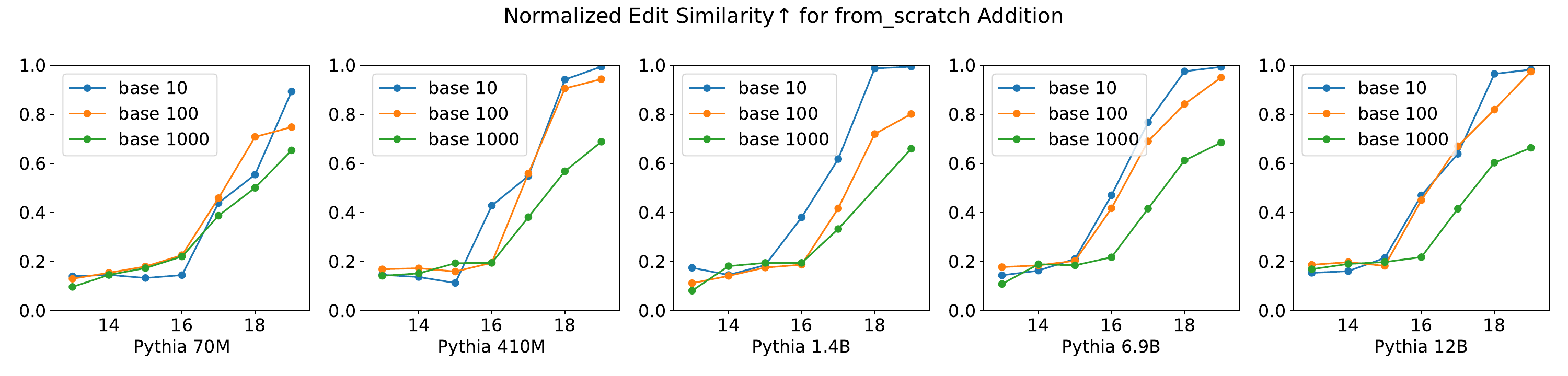}
  \includegraphics[width=\linewidth]{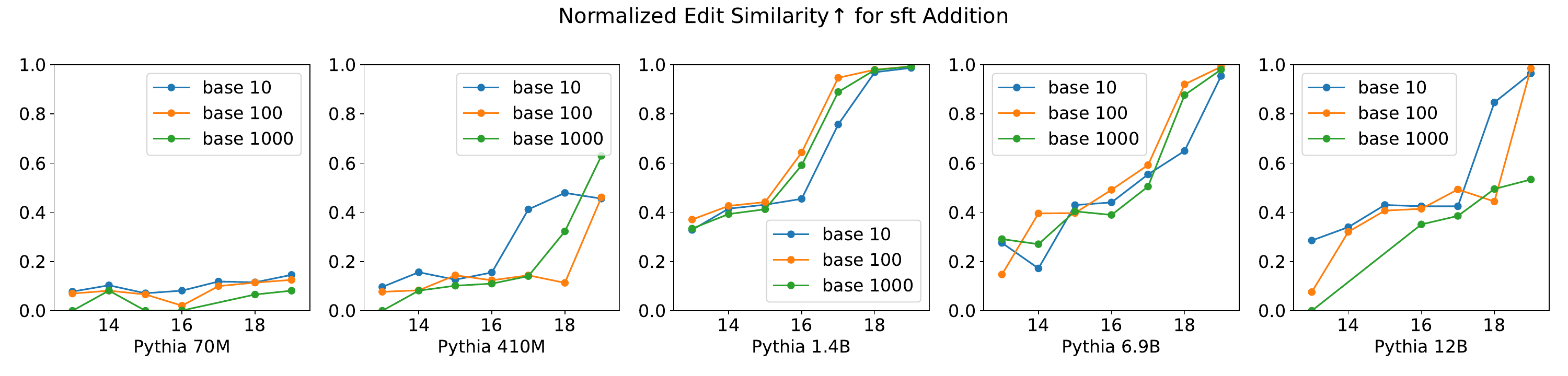}
  \caption {Relative Error (lower is better) and Normalized Edit Similarity (higher is better) for addition operation with different data scales, model parameter sizes, from-scratch or fine-tune, and numeral systems.}
  \label{fig:add_re}
\end{figure*}

To observe a clear trend, we report the following metrics 1) \textbf{relative error} 2) \textbf{exact match accuracy} 3) \textbf{normalized edit similarity}. Of these three metrics, exact match accuracy is the most intuitive, which is a hard match between model-generated tokens and ground truth tokens. Based on our initial experiments, this metric is not informative enough for a range of settings. We thus design two more metrics to reveal the underlying dynamics of our models.

\paragraph{Relative Error}
To generate a metric that is meaningful to practical settings, we calculate \textit{relative error} as $\left\vert \log \frac{o}{g} \right\vert $, where $g$ is the ground truth answer and $o$ is the model output. We then compute the mean of the magnitude difference over all evaluation pairs. This metric is more informative than exact match accuracy since it captures the relative error made by the model. 

Note that this metric has two inductive biases. First, this metric gives more weight to the length differences between model outputs and ground truth answers. Even if the output has a long common sub-sequence with the ground truth, it will still be penalized for not getting the output length right. 

Second, this metric biases towards the accuracy of leading digits. If we make a connection between the numeral system and signal processing, this is equivalent to putting more weight on \textit{low frequency} component of the number (trailing digits change rapidly while leading digits change slowly).
\paragraph{Normalized Edit Similarity}
Since numbers in a numeral system are sequences of tokens, we introduce a generalized and Normalized Edit Similarity metric, which would give credit to partially correct answers based on string similarity.
\textit{Edit Distance} is a powerful metric that can capture substring similarities. We extend this metric to our scenario using the following setup: 

Each number could be represented as a sequence of chars, with each char representing a single digit from $0\sim 9$. We define the \textit{generalized edit distance} as the minimum number of insertions, deletions, and substitutions needed to transform one sequence into another. Suppose that the two sequences are $A = a_1a_2...a_n$ and $B = b_1b_2...b_m$. Let $ed$ be the edit distance between $A$ and $B$. We define the \textit{ normalized edit similarity} as $ned =\frac{\max\limits_{}\left( m, n \right) - ed}{\max\limits_{}\left( m, n \right) }$. 
This metric is normalized into $[0,1]$.

Compared with the \textit{Relative Error} metric, this metric connects more closely to human perception. It prioritizes answers that would have the longest sub-sequences with the ground truth. Since human perception is largely visual for numbers, this metric aligns more with the \textit{visual similarity} between the answer and the ground truth.

Note that \textit{Relative Error} can be somewhat viewed as a revised version of the \textit{Normalized Edit Similarity} we used, where \textit{insert} and \textit{delete} operations are penalized harder, and \textit{replace} operation is reweighted by the magnitude of the difference. 

\begin{figure*}[h] 
  \includegraphics[width=\linewidth]{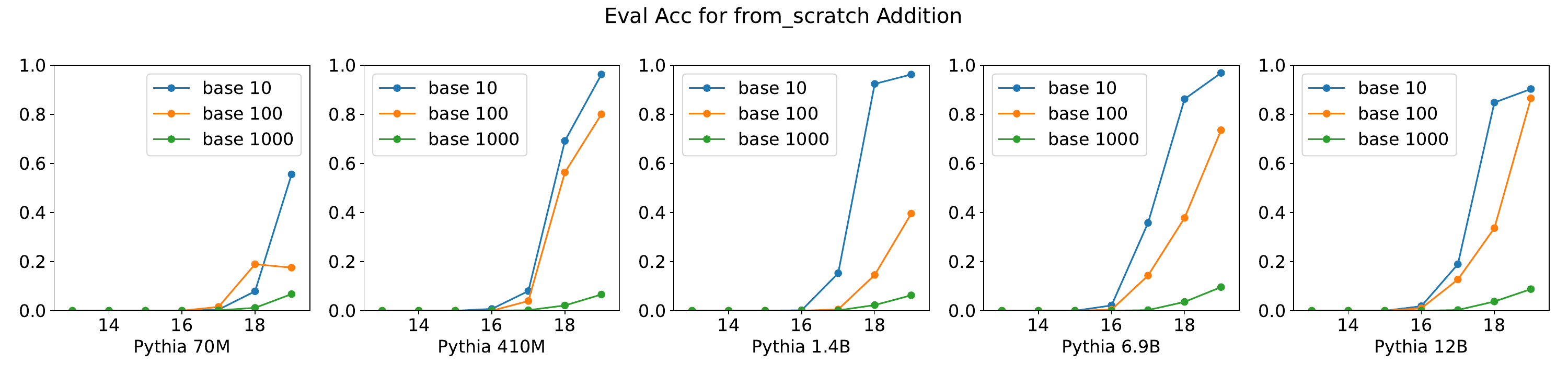}
  \includegraphics[width=\linewidth]{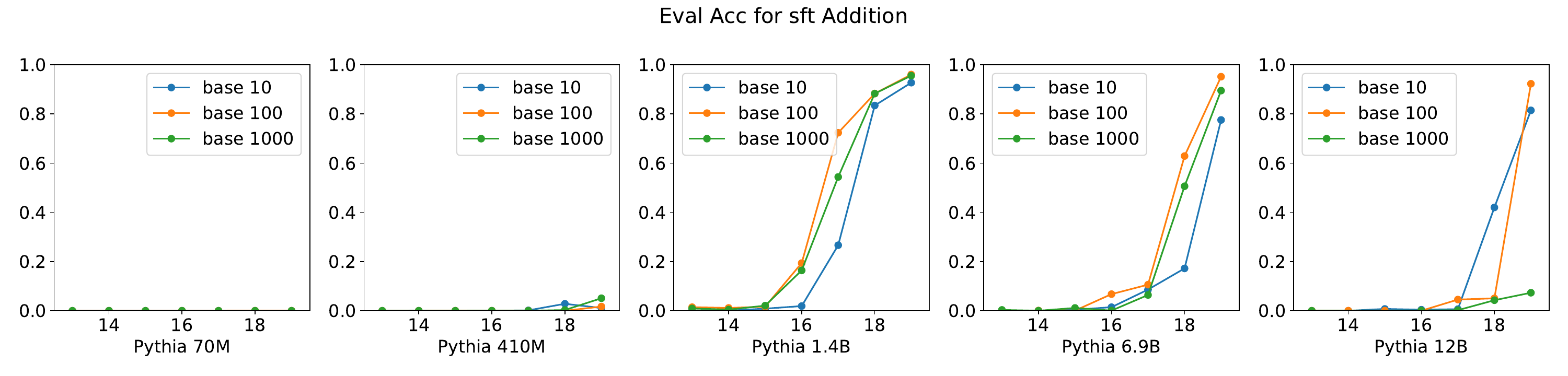}
  \caption {Exact match accuracy for addition operation with different data scales, model parameter sizes, from-scratch or fine-tune, and numeral systems.}
  \label{fig:add}
\end{figure*}

\section{Experiments and Results}
In this section, we present the main results of our experiments that demonstrate the scaling efficiency of a base $10$ system. \footnotemark
\footnotetext{We provide full hyperparameters in Appendix \ref{app:hyperparameters}. We also provide small-scale evaluations in real-world scenarios in Appendix \ref{app:real_world}.}
\subsection{Overall Trends}
For each scenario, our main metrics of interest are \textit{Relative Error} and \textit{Normalized Edit Similarity}. For the addition operation, we also report \textit{Exact Match Accuracy}. However, for multiplication, exact match accuracy is too low such that no information could be gained.

\begin{figure*}[ht] 
  \includegraphics[width=\linewidth]{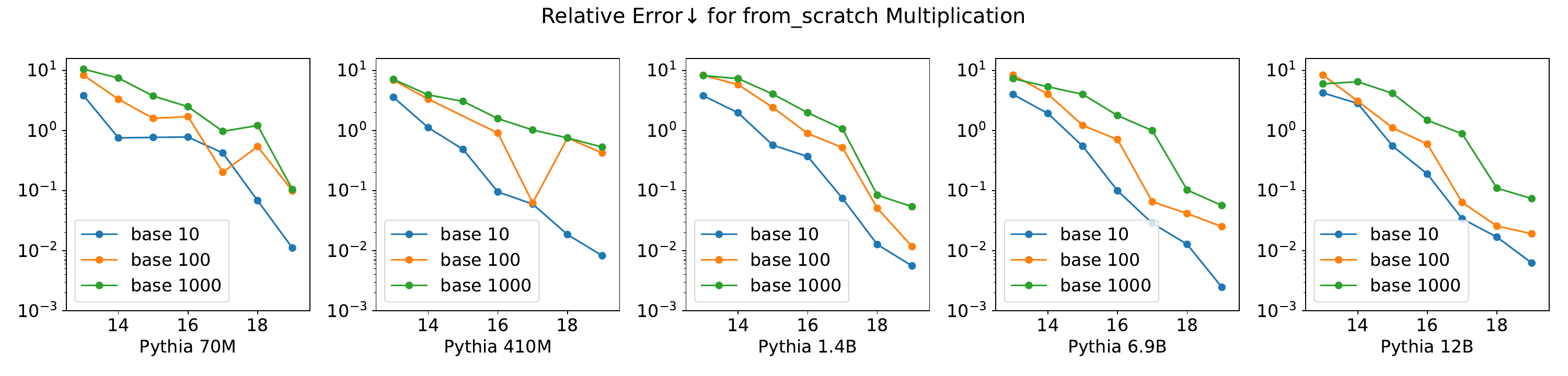}
  \includegraphics[width=\linewidth]{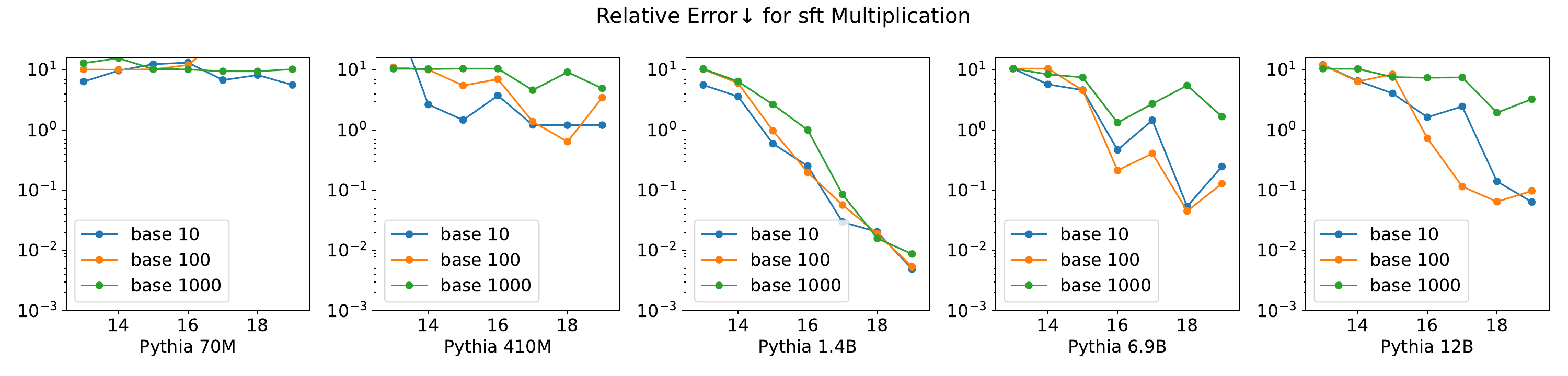}
  \includegraphics[width=\linewidth]{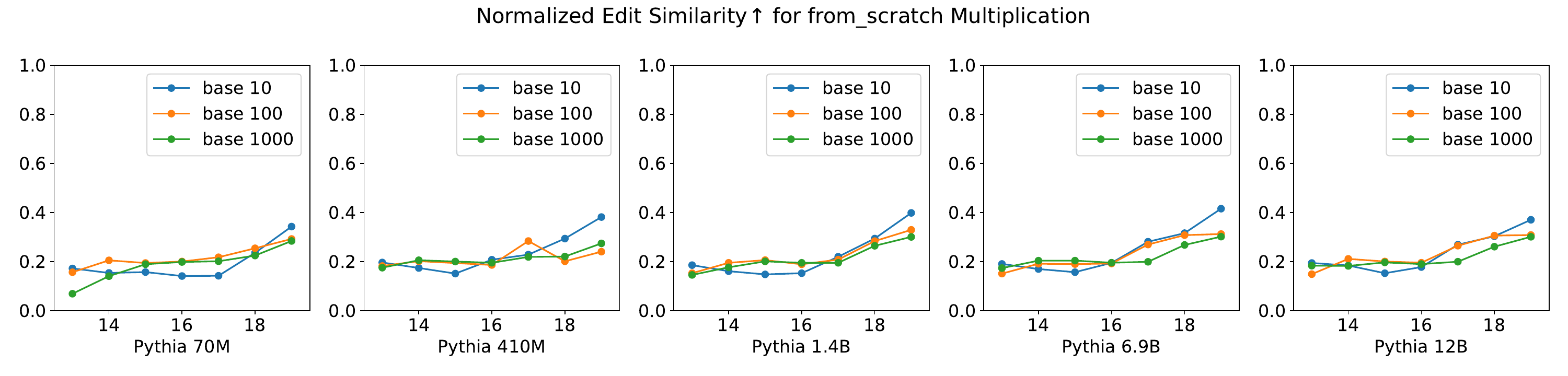}
  \includegraphics[width=\linewidth]{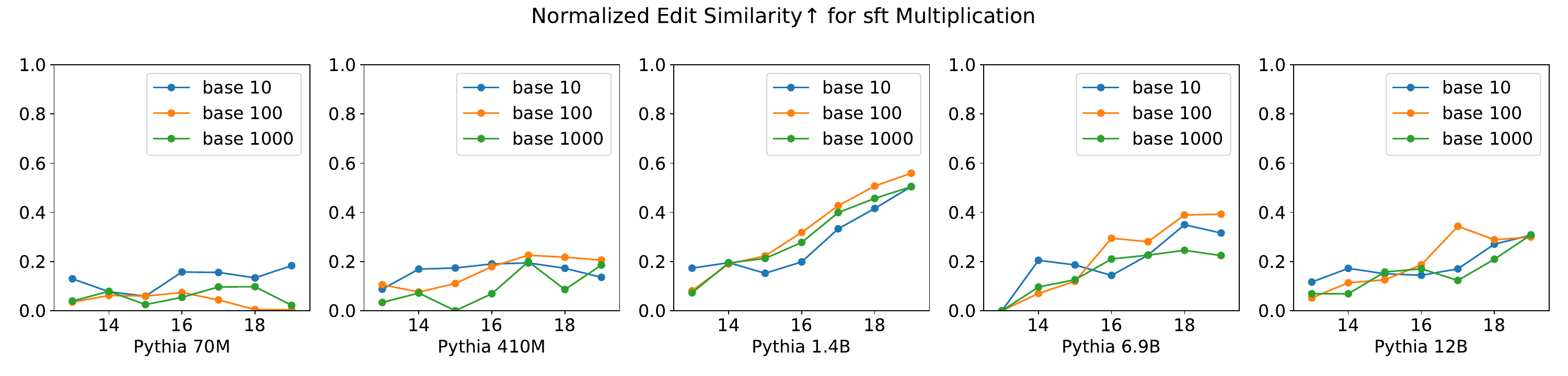}
  \caption {Relative Error and Normalized Edit Similarity for multiplication operation with different data scales, model parameter sizes, from-scratch or fine-tune, and numeral systems.}
  \label{fig:mul_re}
\end{figure*}

Overall, a base $10$ system is \textbf{consistently more data-efficient} than a base $10^2$ or a base $10^3$ system when trained from scratch, as shown in Figure~\ref{fig:add_re} and Figure~\ref{fig:mul_re}. That is, to obtain a certain performance, a base $10$ system would need data of a smaller scale to achieve it. 

We highlight the fact that \textbf{fine-tuning} scenarios do \textbf{not} instead favor base 100 and base 1000. During pre-training, most tokenizers lean towards combining consecutive digits, which would have favored base $10^2$ and base $10^3$ over base $10$. Considering this, the decent performances of base $10$ fine-tuned models further corroborate the superiority of the base $10$ system.

It is worth noting that the differences in data efficiency do not diminish just as we scale up model size. We do observe a saturation trend for addition when we put in more training data. However, for multiplication, the superiority of a base $10$ system gets \textbf{more pronounced} as data scales up.

\subsection{In-domain Interpolation Evaluation}
First, we test whether our trained models could interpolate between the points identically sampled from the same training distribution (i.e. whether our models could generalize in-domain).
\subsubsection{Addition}
In Figure \ref{fig:add_re}, we showcase the scaling behavior for addition. We first focus on \textit{from-scratch} scenario. We can observe a clear trend that base $10$ is consistently better than base $10^2$ and base $10^3$ for both metrics, which is a strong affirmation of our claim. A base $10$ system is at least of a constant magnitude more data efficient than base $10^2$ and base $10^3$ systems, and this trend does not diminish as the model size gets larger.

For fine-tuning experiments, the difference between numeral systems is less profound. 
Pythia is pre-trained on tokenization with base $10^3$, which weakens the advantages of base $10$.
A base $10$ system is at least on par with base $10^2$ or base $10^3$, as we do not observe an exaggerated performance difference as we scale up data. 

\subsubsection{Multiplication}
We plot the \textit{Normalized Edit Similarity} for multiplication in Figure \ref{fig:mul_re}. We can also conclude that the base $10$ number system is consistently more data-efficient than base $10^2$ and base $10^3$. The trend is consistent for both \textit{from-scratch} and \textit{supervised fine-tuning} settings. 

The superiority of a base 10 system is more pronounced and more consistent under the multiplication setting. First, for Relative Error of models trained from scratch, the advantage of a base 10 system is more perceivable than the addition setting. For the Normalized Edit Similarity metric, we observe a trend where the data efficiency of a base 10 system gains more advantage at large data scales. We relate this phenomenon to the differences between Figure \ref{fig:mul_matrix} and Figure \ref{fig:add_matrix} in the Appendix. As shown, addition is a much simpler task when compared with multiplication. For a large range of operand length pairs, the exact match accuracy remains zero. 
Our hypothesis is that the sample efficiency of a base 10 system against base 100 and base 1000 systems is magnified by the difficulty of the task.

\subsection{Out-of-domain Extrapolation Evaluation}
We have tested whether our models could generalize in-domain. An equally important question is whether our models could \textbf{extrapolate} to unseen data points, especially in terms of \textbf{length}.

During training distribution generation, we only consider numbers that are less than $10^{11}$. Therefore, we generate cases where one operand lies in the range of $10^{11}\sim10^{16}-1$, and the other operand ranges from $1\sim10^{16}-1$.
\begin{figure}[h] 
  \includegraphics[width=\columnwidth]{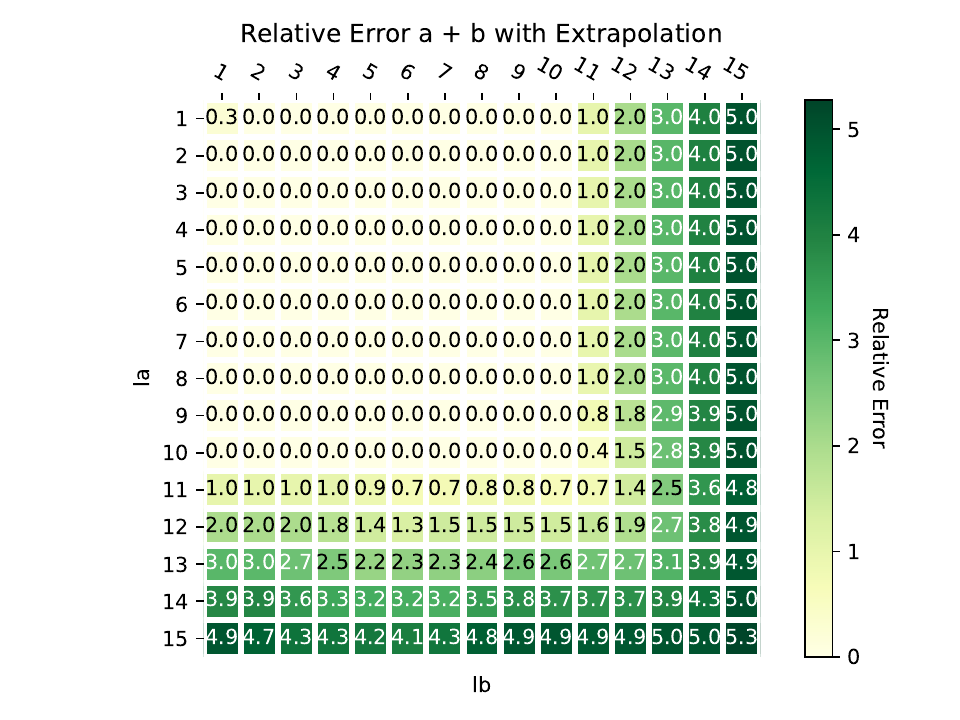}
  \caption {Relative Error Matrix for Extrapolation Behavior Analysis. The results are obtained using a 1.4B model fine-tuned on $2^{19}$ training samples.}
  \label{fig:add_extrapo_matrix}
\end{figure}
To this end, we perform 12 sets of case study experiments. Here we list the complete configurations: 1) numeral system: base $10$, $10^2$, $10^3$ 2) data scale: $2^{19}$ 3) model size: 6.9B 4) if from-scratch: True, False 5) operations: addition, multiplication. We discover intriguing extrapolation behavior that could shed light on the mechanisms that the models have learned.

\begin{table*}[ht]
    \centering
    \small
    \begin{tabular}{llllll}
        \toprule
        Model & Pattern & a & b & $a + b$ & Model Output \\
        \midrule
        SFT-Base $10$ &w/o carry &8318686348,0 &  3 & 8318686348,3 & 83186863\textcolor{blue}{51} \\
        &carry success &3968299531,8 & 2& 3968299532,0& 396829953\textcolor{blue}{4} \\
        \midrule
        SFT-Base $10^2$ & w/o carry &7 34 76 64 43, 03 & 3 & 7 34 76 64 43, 06 &  7 34 76 64 \textcolor{blue}{46} \\
        & misaligned& 1 63 47 53 10, 81 & 2 & 1 63 47 53 10, 83 & 1 63 47 53 \textcolor{blue}{12}, 83 \\
        & carry failure&7 28 37 46 59, 47 & 94 & 7 28 37 46 60, 41 & 7 28 37 \textcolor{blue}{47 53} \\
        \midrule
        SFT-Base $10^3$ & w/o carry & 2 929 747 175, 022 & 9 & 2 929 747 175, 031 &  2 929 747 \textcolor{blue}{184} \\
        & misaligned& 8 748 392 297, 087 & 2 & 8 748 392 297, 089 & 8 748 392 \textcolor{blue}{299}, 089 \\
        & carry failure& 8 172 938 472, 837 & 494 & 8 172 938 473, 331 & 8 172 938 \textcolor{blue}{966} \\
        \bottomrule
    \end{tabular}
    \caption{Representative Cases for Addition Extrapolation. We add a comma to denote the maximum token length of a single number that the model has seen during training.}
    \label{tab:case_study}
\end{table*}

\begin{table*}[h]
    \centering
    \small
    \begin{tabular}{llllll}
        \toprule
        Model& a& b& $a\times b$& Model Output\\
        \midrule
        SFT-Base $10$&9298574444, 7&6&5579144666,82&5579144666,82\\
        SFT-Base $10^2$&3 44 97 17 48, 09&8&27 59 77 39 84, 72&27 59 77 39 84, 72\\
        SFT-Base $10^3$&18 419 335, 384&4&73 677 341, 536&73 677 341, 536\\
        \midrule
        Scratch-Base $10$&44527557923&8&356220463384&35\textcolor{blue}{88888999}84\\
        Scratch-Base $10^2$&2 45 57 14 10, 66&8&19 64 57 12 85, 28&\textcolor{blue}{20} \textcolor{blue}{10} \textcolor{blue}{88} 12 \textcolor{blue}{12}, \textcolor{blue}{48}\\
        Scratch-Base $10^3$&17 709 751, 495&5&88 548 757, 475&\textcolor{blue}{87} \textcolor{blue}{229} \textcolor{blue}{700}, \textcolor{blue}{075}\\
        \bottomrule
    \end{tabular}
    \caption{Representative Cases for Multiplication Extrapolation. We list successful cases for fine-tuned models (i.e. SFT) and showcase the failure of from-scratch models.}
    \label{tab:case_mul}
\end{table*}

\subsubsection{Addition}
Note that although addition is an easy task to train, the models have only seen numbers less than $10^{11}$. We leverage addition pairs of length $la$ and $lb$, where at least one of $la$ or $lb$ is greater than 10. To illustrate how the performance of our model decays, we plot the \textit{Relative Error} matrices where one operand length is in $[1, 5]$ and the other in $[11, 15]$ in Figure \ref{fig:add_extrapo_matrix}. The results are obtained using a 1.4B fine-tuned model trained on $2^{19}$ training samples. For each pair of $la \times lb$, we randomly generate 100 samples, which results in a total of 5000 samples. Of all such samples, the extrapolation exact match accuracy is $0.0$. 

Yet, the models do not collapse completely on out-of-domain length distribution. We conduct case studies in Table~\ref{tab:case_study}.
Our first discovery is that there is a consistent behavior of \textit{Truncated Addition} across all numeral systems of fine-tuned models. Our second observation is that fine-tuned models are much better at aligning the tokens involved in extrapolated addition, as compared with models trained from-scratch.\\
\noindent \textbf{Truncated Addition Extrapolates} \quad While we are manually inspecting the extrapolation behavior of fine-tuned models, we discover consistently that models would try to perform the addition \textbf{truncating} the tokens that exceed training length it has seen. We elaborate on this behavior under two configurations. 

For illustrative purposes, we add a comma to denote the max training length position the models have seen. First, take for example a model trained on a base $10$ system, a fine-tuned model is given input $a=8318686348,0$ and $b=3$, where the token representation of $a$ is 11. Only the first 10 digits of $a$ would participate in addition, yielding a result of $8318686348 + 3 = 8318686351$. A fine-tuned model trained on the base $10^2$ system displays very similar results. Given $a=734766443,03$ and $b=3$, the model performs $734766643 + 3 = 734766446$, ignoring two trailing digits, which is equivalent to ignoring the last token under base $10^2$.
The phenomenon of truncated addition is hardly observed on models trained from-scratch. The main obstacle could arise from the inability to \textbf{align} corresponding tokens with unseen token lengths. For example, a base $10$ model trained from-scratch would calculate $2635078980,7 + 1 = 263507\textcolor{blue}{9091}$, where the 1 seems to have been added to multiple positions. This could also indicate that fine-tuned models have learned to utilize positional information better.\\
\noindent \textbf{Base $10$ Carry Extrapolates} \quad While we attempted to explain extrapolation behavior using \textit{truncated addition}, we noticed some outliers where the answer is only 1 absolute value larger than the \textit{truncated addition} result. Manual inspection quickly reveals that the models generate carry for out-of-distribution positions. For a base $10$ fine-tuned model, $3968299531,8 + 2 = 396829953\textcolor{blue}{4}\ (= 3968299531 + 2 + 1)$, where a carry has been generated because $8 + 2  = 10$. Note that the carry is not generated by aligning the ones digit since $1 + 2 = 3 < 10$, which is an ablation showcasing that calculating carry exhibits extrapolation behavior.

\noindent \textbf{Tokens Generalize, Length Does Not} \quad For a base $10^3$ system, two kinds of behavior have been observed. Before we describe the behaviors, we restate our experiment settings. Our training distribution only contains numbers that are less than $10^{11}$ under the base $10$ system. This creates two scenarios for extrapolation experiments of a model trained with base $10^3$. 1) both operands are less than $10^{13}$ 2) at least one of the operands is no smaller than $10^{13}$. For 1), although the model has not seen any data points within the range of $10^{11}\sim10^{13}-1$, the length of both operands does not exceed 4, which has been trained. For 2), the length of at least one operand has not been seen during training at all. 

Out of a sample size of 100 for each $la \times lb$ pair, a base $10^3$ fine-tuned model could achieve 90\% exact match accuracy with $la=11, lb\in [1, 8]$. However, accuracy \textbf{quickly drops to 0} if one of the operands has a token length greater than 4 under the base $10^3$ system.

\subsubsection{Multiplication}
Different from addition, there is at least one successful example of extrapolation of operand length for fine-tuned models of all number systems shown in Table~\ref{tab:case_mul}. Yet, the exact match accuracy on the extrapolation set of models tuned from-scratch is consistently zero. Moreover, a closer look at the generated results showcases that the model is only able to correctly generate the starting tokens and ending tokens of the answer, with gibberish and repetitive tokens in the middle. 


\section{Conclusion}
In this paper, we study the selection of a numeral system for large language models. We compare the data-scaling efficiencies of base 10, 100, and 1000 systems. Through carefully designed experiments, we showcase the superiority of the base 10 system. 

We offer an analysis of the extrapolation behavior of trained models on addition and multiplication. We reveal calculation patterns that successfully extrapolate, such as carry generation in addition and magnitude estimation in multiplication. Our work sheds light on tokenization designs and the mechanisms that models have learnt for arithmetic tasks.

\section*{Limitations}
Scaling behavior analysis requires a huge amount of computational resources. Limited by this factor, we have not performed a thorough grid search for hyperparameters of every setting. It is possible that for every configuration that is of interest, we should use a unique set of hyperparameters to achieve optimal performance. 
In our experiments, we have witnessed instability issues regarding some data points where the training loss seemingly collapses. It is possible that such issues arose because of suboptimal hyperparameter choices.
Research on numerical operation has few potential risks.


\bibliography{custom}

\clearpage
\appendix

\newpage
\section{Discussion on Multi-digit Tokenization} \label{app:tokenization}
In this paper, we showcase the superior data efficiency of a base $10$ system. Yet, in Table \ref{tab:segmentation}, popular models such as Llama3 and GPT-4o still adhere to multi-digit tokenization for numbers. We provide a discussion about this phenomenon here. 

From our perspective, this design choice relates to user experience and cost management. During inference, using a larger numeral system reduces the total number of tokens, leading to shorter input and output lengths. This reduction boosts token throughput (due to smaller kv cache size) and increases the number of queries per second (because output is shorter), significantly improving user experience and reducing training/inference costs.

One plausible assumption is that for a base $100$ or base $1000$ system, one can devote more computer into training to trade-off for better inference experience. It is likely that the efficiency for training and inference could not be improved simultaneously. 
\section{Discussion on Task Selection}\label{app:task_choice}
We provide a detailed reasoning for choosing \textit{direct} \textbf{addition} and \textbf{multiplication} as our tasks (or arithmetic operations) to investigate.

First, we strengthen that our setting targets \textit{direct} calculation of addition and multiplication. For example, the model is directly prompted with $13 + 5 =$, (or $13 \times 5 =$). The model is expected to directly output $18$, (or $65$). 

We noticed that the accuracy of direct calculation upper bounds model's ability to carry out calculations \textbf{in context}. Under such settings, the model need to perform calculations mixing natural language and numbers. For example, the model might output "To solve the problem, we need to calculate the product of $13 \times 5$, which is 65". This motivates us to investigate direct calculations first.

Second, we address why we only presented results on \textbf{addition} and \textbf{multiplication}. In our initial small-scale experiments, we found out two things. First, for division, pre-trained models have trouble ending their outputs. Second, for subtraction, we found trends that are similar to addition. We also argue that arithmetic tasks have shared attributes with addition and multiplication: 1) For each task, a token-level operation should be learned (eg. adding single-digit numbers). 2) For each task, the model would need to discern between the tokens.

We would like to state that our work has a potentially broader impact. Our generalized finding is that having fewer states could enhance sequential modeling. For other planning tasks, such as robotics or theorem proving, this conclusion may also hold true.
\section{Connection with Scaling Law}\label{app:scaling_law}
We intentionally avoid using the term \textbf{Scaling Law}. Generally, researchers fit a law to predict the performance of models when scaling up compute. However, for addition, we already observe saturation in Figure \ref{fig:add_re}. It is hard to accurately fit a law for this curve. 

Moreover, we are not interested in \textbf{predicting} performances. We are inspecting the difference in \textbf{data efficiency} as we scale up compute. Therefore, observing that a base $10$ system is more efficient is eloquent enough for our purpose.

We also find that our performances do \textbf{not} improve significantly with \textbf{model sizes}. We have noticed that the 70M and 410M versions of Pythia models are particularly hard to train. We do not dive into this technical detail. 
\section{Metric Matrices for Length Pairs}
\begin{figure}[t] 
  \includegraphics[width=\columnwidth]{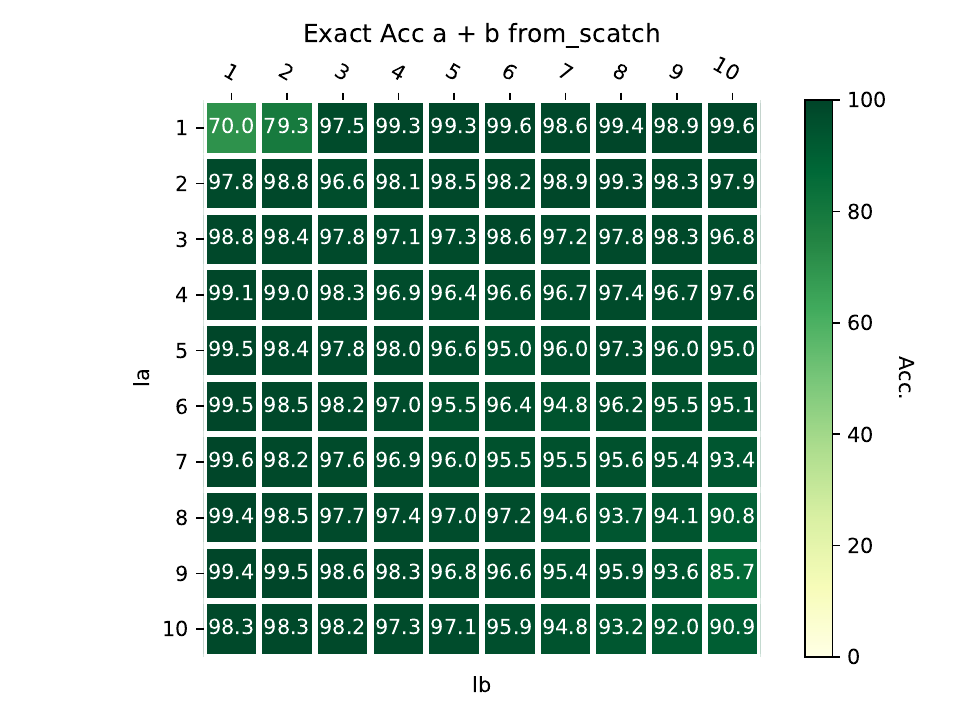}
  \includegraphics[width=\columnwidth]{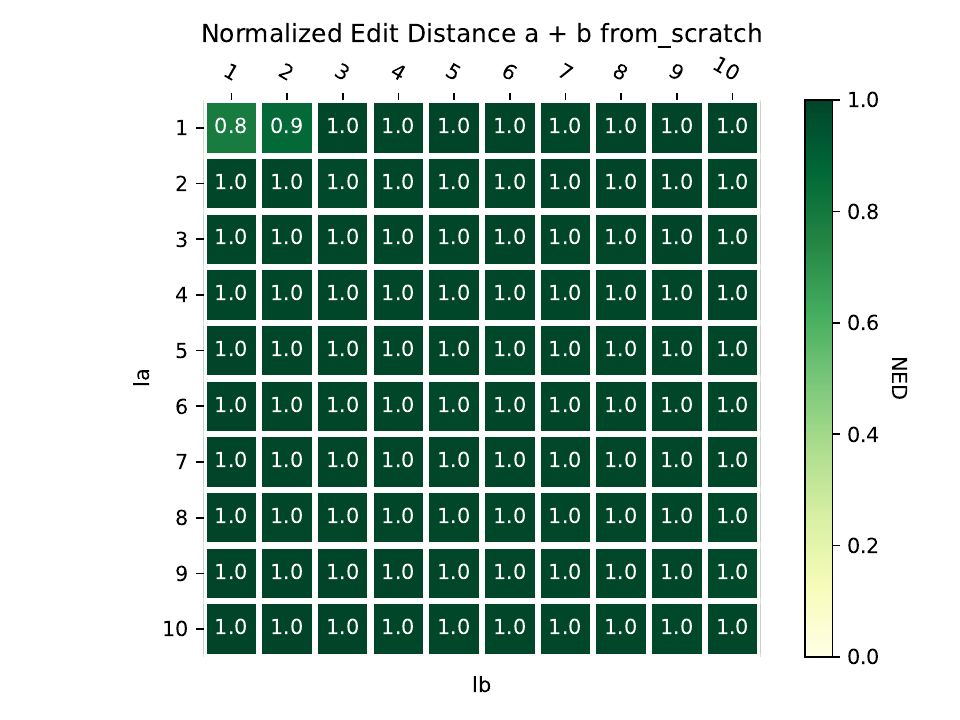}
  \caption {Exact Match Accuracy and Normalized Edit Similarity Matrices for Addition Eval Set. The results are obtained using a 1.4B model trained from scratch on $2^{19}$ samples.}
  \label{fig:add_matrix}
\end{figure}

\begin{figure}[t] 
  \includegraphics[width=\columnwidth]{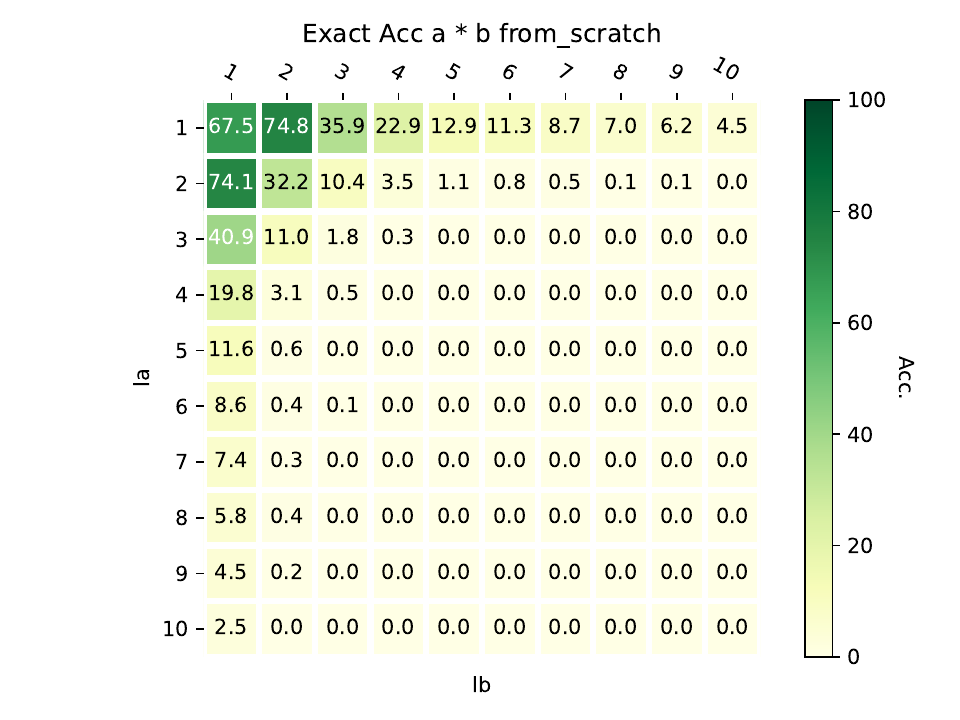}
  \includegraphics[width=\columnwidth]{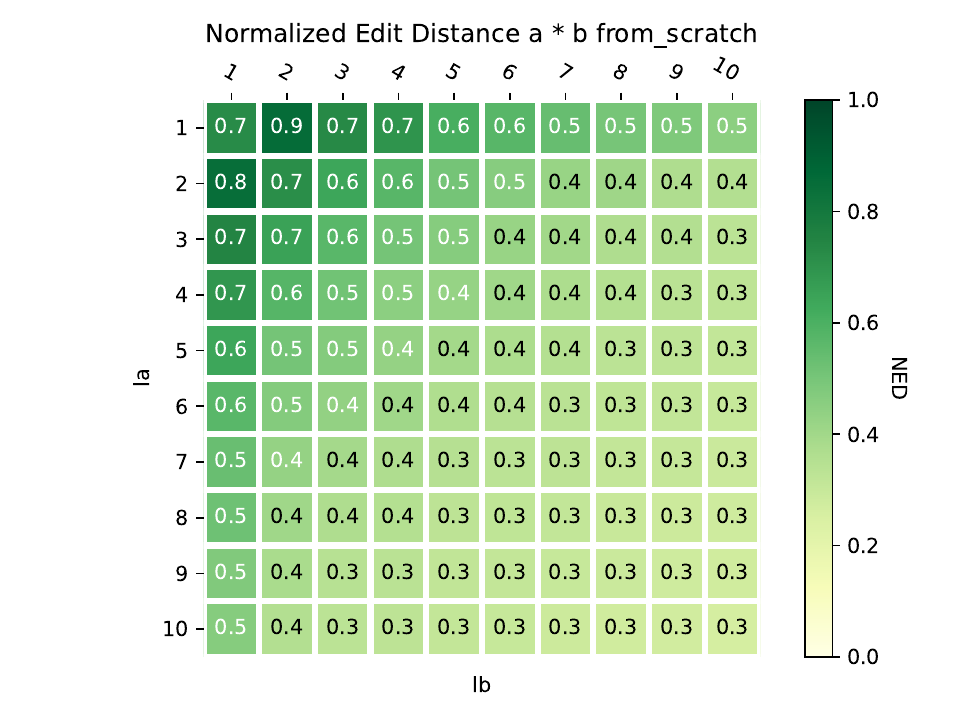}
  \caption {Exact Match Accuracy and Normalized Edit Similarity Matrices for Multiplication Eval Set. The results are obtained using a 1.4B model trained from scratch on $2^{19}$ samples.}
  \label{fig:mul_matrix}
\end{figure}

We take a 1.4B model trained from-scratch on addition and multiplication as an exemplar and plot matrices for both Exact Match Accuracy and Normalized Edit Similarity with respect to each pair of input lengths in Figure~\ref{fig:add_matrix} and Figure~\ref{fig:mul_matrix}.

\section{Overfitting Analysis}
\begin{figure*}[th] 
  \includegraphics[width=\linewidth]{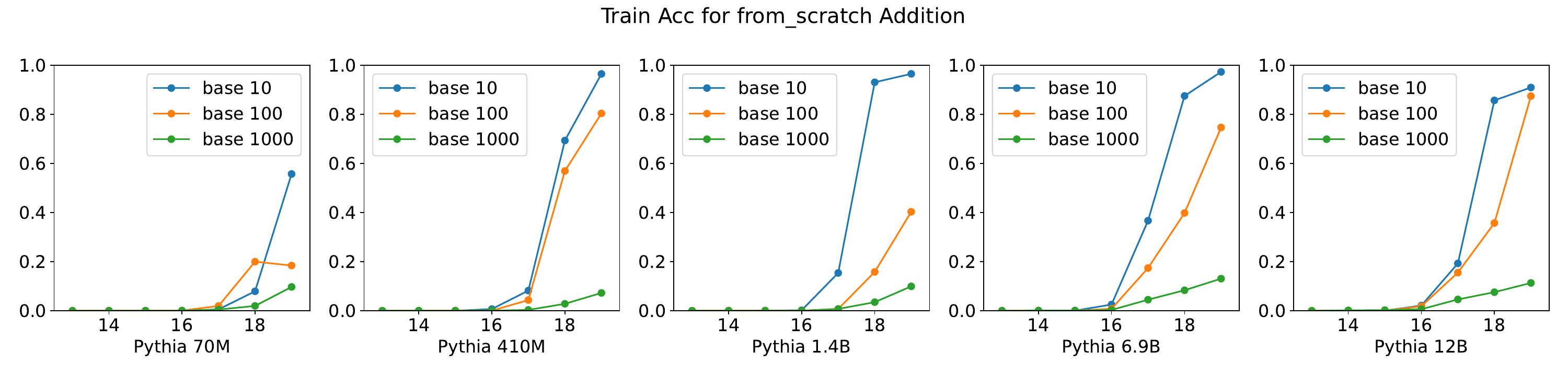}
  \includegraphics[width=\linewidth]{figures/add_acc-fs.pdf}
  \caption {Exact Match Accuracy on the training set versus the eval set for addition operation with different models trained from scratch, on different data scale and numeral systems.}
  \label{fig:overfit}
\end{figure*}

Alongside our main results, we also perform ablation studies on overfitting under addition settings, since the accuracy quickly saturates to 100.0\%. First of all, we subsample a portion of our training set to forward through the model. We attempt to sample 1000 examples for each $la \times lb$ pair in $10\times 10$. If the total number of training pairs is smaller, we take all training pairs for $la \times lb$.

Generally, for all the metrics of interest, we observe nearly identical performance on our training and evaluation set. One example of accuracy on addition is shown in Figure \ref{fig:overfit}. Furthermore, since our evaluation set is non-overlapping with the training set, it would be safe to conclude that no overfitting phenomenon has been observed.

\section{Hyper-parameters}
\begin{table*}[th]
    \centering
    \begin{tabular}{|c c c c c c|}
    \toprule
    \textbf{Model Version} & \textbf{From Scratch} & \textbf{Learning Rate} & \textbf{Max Epochs} & \textbf{Log10} & \textbf{Data Scale} \\ \midrule
    pythia\_70m   & True  & 2e-4  & 10  & 1-3  & 13-19 \\
    pythia\_70m   & False & 2e-5  & 2   & 1-3  & 13-19 \\
    pythia\_410m  & True  & 2e-4  & 10  & 1-3  & 13-19 \\
    pythia\_410m  & False & 3e-5  & 2   & 1-3  & 13-19 \\
    pythia\_1\_4b & True  & 2e-5  & 10  & 1-3  & 13-19 \\
    pythia\_1\_4b & False & 2e-5  & 2   & 1-3  & 13-19 \\
    pythia\_6\_9b & True  & 2e-5  & 10  & 1-3  & 13-19 \\
    pythia\_6\_9b & False & 2e-5  & 2   & 1-3  & 13-19 \\
    pythia\_12b   & True  & 2e-5  & 10  & 1-3  & 13-19 \\
    pythia\_12b   & False & 2e-5  & 2   & 1-3  & 13-19 \\
    \bottomrule
    \end{tabular}
    \caption{Hyperparameters for Multiplication}
    \label{apptab:hp_mul}
\end{table*}
We briefly discuss the hyperparameter search process that we have gone through for each configuration. Based on our initial experiments, models exhibit nearly identical behavior in both the training set and the evaluation set. We therefore use training set metrics for hyperparameter selection. 

Then, we first fix the learning rate magnitude and sweep for training epochs. We observed that fine-tuned models are insensitive for training epochs, while the model training from-scratch consistently improves with more epochs. We choose epochs where the performances of models begin to plateau. Generally, epoch performance trends only depend on the training setting (i.e. fine-tuning or from-scratch).
Fixing the training epoch, we perform a grid search over learning rate magnitudes $\{2e-3,2e-4,2e-5,2e-6,2e-7\}$ for each configuration. Generally, we found that 70M and 410M models favor a larger learning rate of $2e-4$ while models larger than $1.4B$ use $2e-5$. There is no significant difference between fine-tuning learning rates and from-scratch learning rates.
We provide our full hyperparameters in Table \ref{apptab:hp_mul}. The hyperparameters for addition is the same as multiplication, except that we used $2e-5$ for fine-tuning 410M Pythia.

To speed up training, we pack sequences to a maximum length of 2048, therefore fixing the batch size.
All experiments are trained using 8xA100 Nvidia GPUs.
\label{app:hyperparameters}

\section{Performance Differences in Real-World Models}\label{app:real_world}
\begin{figure*}[th] 
  \includegraphics[width=\linewidth]{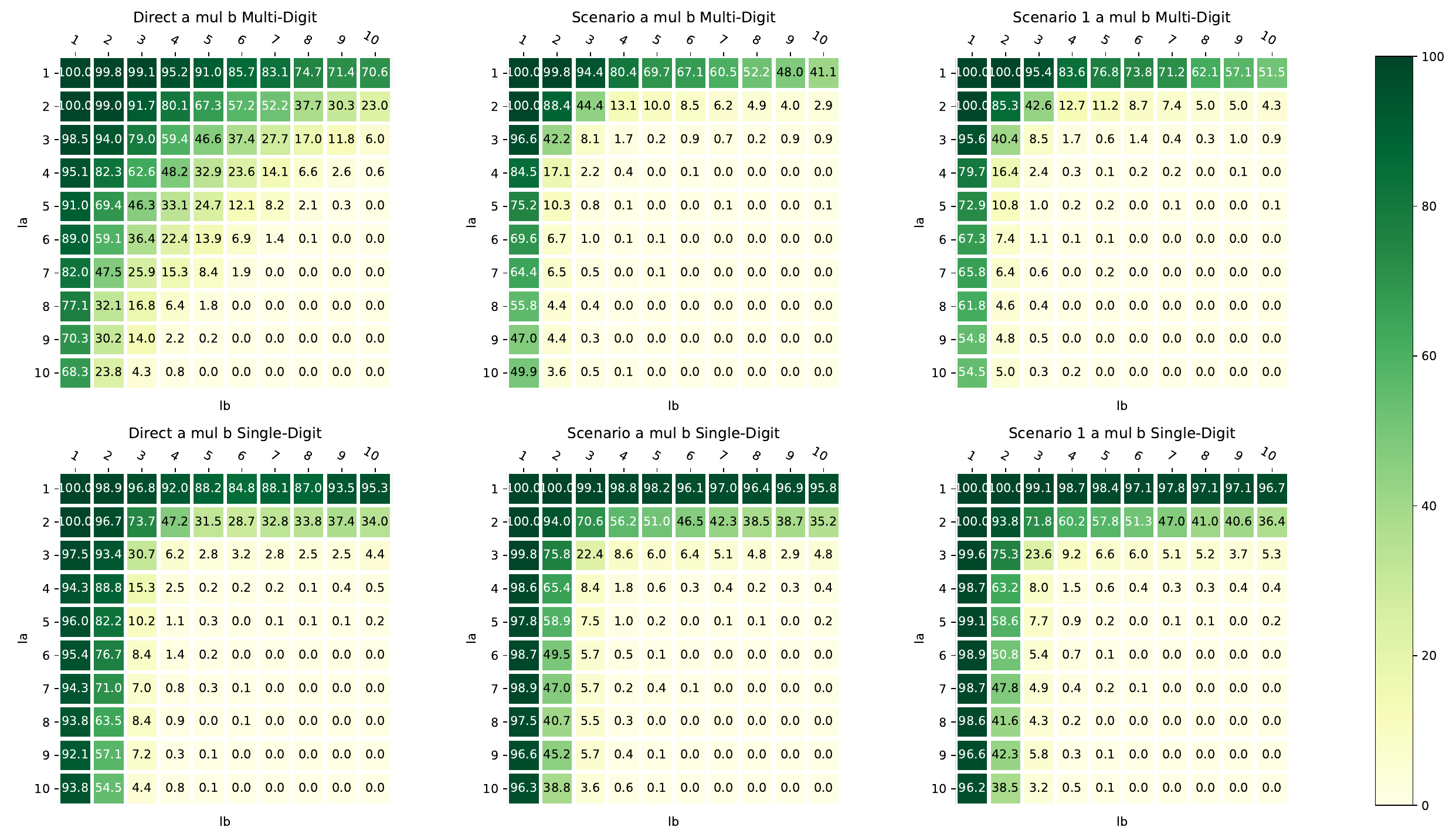}
  \caption {Performances of difference tokenization schemes in three real-world scenarios: 1) directly calculating $a \times b$, 2) solving an natural language application problem that involves solving $a \times b$, and 3) solving a variant application problem that also involves solving $a \times b$.}
  \label{appfig:real_world_acc_diff}
\end{figure*}

One reasonable question to ask is whether our finding is useful in training real-world scenario math models. We present our results in Figure \ref{appfig:real_world_acc_diff}. We test two versions of models: 1) \textit{Multi-Digit Tokenization}, and 2) \textit{Single-Digit Tokenization}. We test our conclusion under three scenarios: 1) directly calculating $a \times b$, 2) solving an natural language application problem that involves solving $a \times b$, and 3) solving a variant application problem that also involves solving $a \times b$. We generate test data using a program and extract the answer using heuristics. We report accuracies using hard match.

Two observations could be made. 1) \textit{Single-Digit Tokenization} consistently outperforms \textit{Multi-Digit Tokenization}, and 2) the accuracy of directly calculating $a\times b$ gives a reasonable upper-bound of the model's performance when solving an application problem that involves multiplication.
\section{Results for Conventional Relative Error}
\begin{figure*}[th] 
  \includegraphics[width=\linewidth]{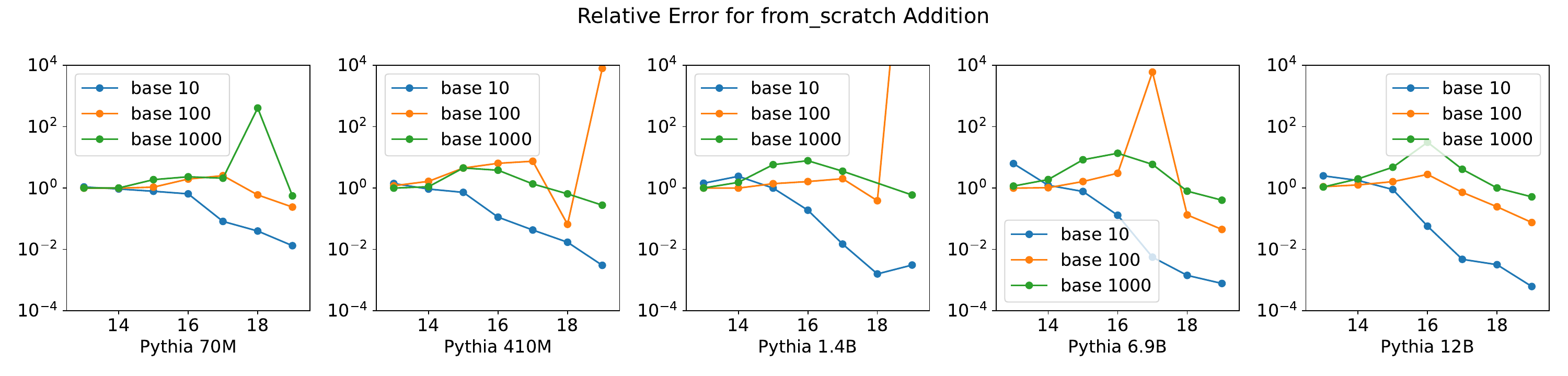}
  \includegraphics[width=\linewidth]{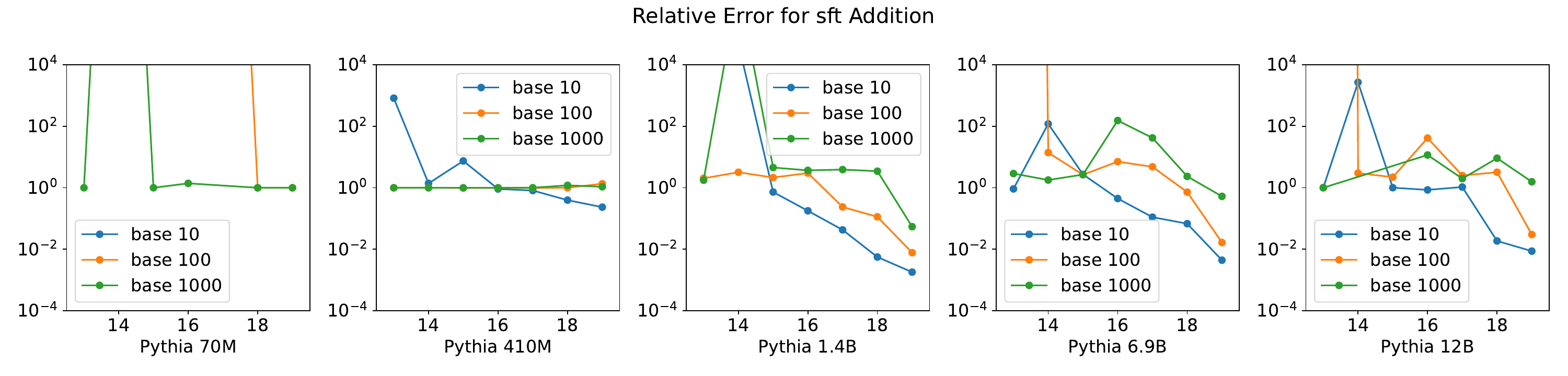}
    \includegraphics[width=\linewidth]{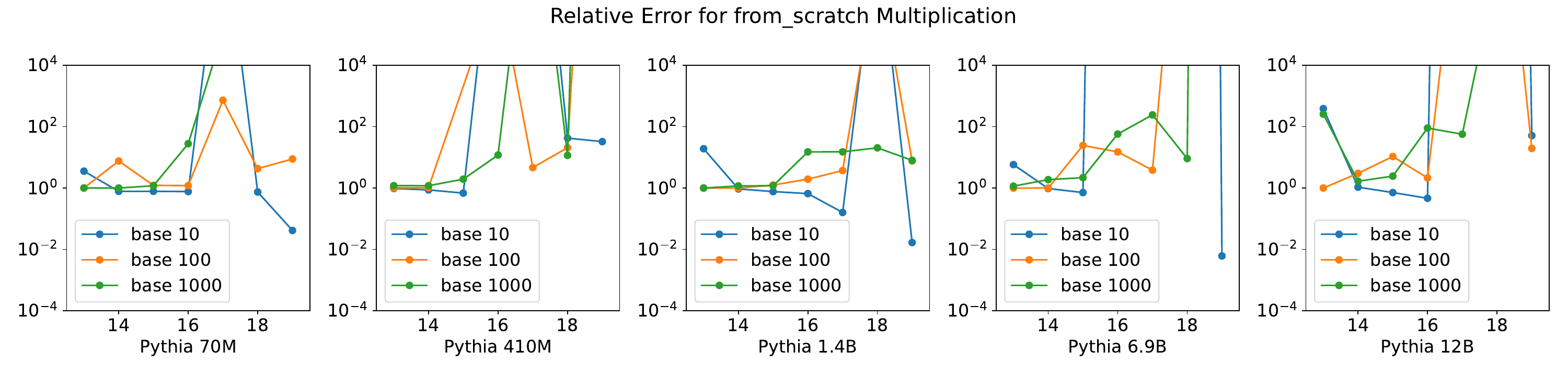}
  \includegraphics[width=\linewidth]{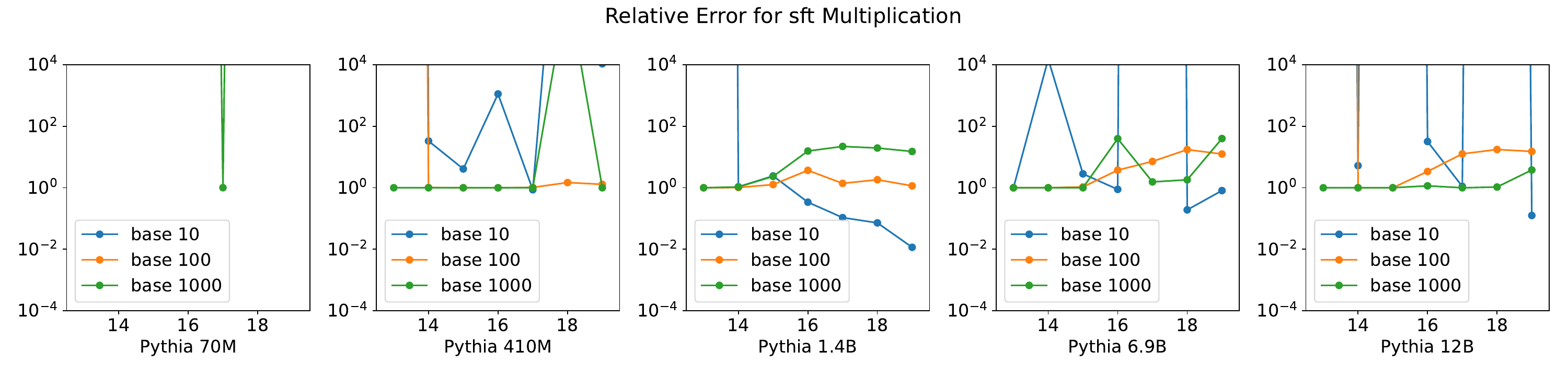}
  \caption {Relative Error results computed using conventional $re=\frac{\left|o-g\right|}{g}$. We see significant issues of instability as shown in the figures.}
  \label{appfig:re}
\end{figure*}

One might argue that the \textit{Relative Error} metric that we used in the paper diverges from conventional calculation. Furthermore, it has similarities with the \textit{Normalized Edit Similarity} metric that we used. Here, we plot the results where $RE=\frac{\left|o-g\right|}{g}$ in Figure \ref{appfig:re}. 

For Addition, the advantage of a base 10 system is magnified using conventional calculations. For Multiplication, we witness significant issues of instability, hindering the discovery of underlying insights.
\end{document}